\documentclass{article}
\pdfoutput=1
\PassOptionsToPackage{numbers, compress}{natbib}

\usepackage[final]{neurips_2020}

\usepackage[utf8]{inputenc} 
\usepackage[T1]{fontenc}    
\usepackage{hyperref}
\hypersetup{colorlinks,citecolor=blue,urlcolor=red,linkcolor=red,bookmarks=false}
\usepackage{url}            
\usepackage{booktabs}       
\usepackage{amsfonts}       
\usepackage{nicefrac}       
\usepackage{microtype}      

\usepackage{times}  
\usepackage{amsmath}
\usepackage{graphicx}  

\usepackage{amssymb}
\usepackage{authblk} %
\usepackage{amsthm}
\usepackage{subcaption}
\usepackage{lipsum}
\usepackage{tikz}
\usepackage{wrapfig}
\usepackage{epstopdf}
\usepackage{enumitem}
\usepackage[outercaption]{sidecap}    
\usepackage{multirow}
\usepackage{mwe}
\usepackage{tikz}
\usetikzlibrary{positioning}
\usepackage{algorithm}
\usepackage{algorithmic}
\usepackage{comment}
\usetikzlibrary{shapes,snakes}

\usetikzlibrary{fit,positioning}

\newcommand\independent{\protect\mathpalette{\protect\independenT}{\perp}}
\def\independenT#1#2{\mathrel{\rlap{$#1#2$}\mkern2mu{#1#2}}}

\newcommand\blfootnote[1]{%
  \begingroup
  \renewcommand\thefootnote{}\footnote{#1}%
  \addtocounter{footnote}{-1}%
  \endgroup
}

\newtheorem{Theorem}{Theorem}

\title{Sample-Efficient Reinforcement Learning via Counterfactual-Based Data Augmentation}

\author{Chaochao Lu$^{*1, 3}$, Biwei Huang$^{*2}$, Ke Wang$^5$, Jo{\'s}e Miguel Hern{\'a}ndez-Lobato$^{1,4}$, Kun Zhang$^2$, and Bernhard Sch{\"o}lkopf$^3$}

\begin{document}

\maketitle

\begin{abstract}
Reinforcement learning (RL) algorithms usually require a substantial amount of interaction data and perform well only for specific tasks in a fixed environment. In some scenarios such as healthcare, however, usually only few records are available for each patient, and patients may show different responses to the same treatment, impeding the application of current RL algorithms to learn optimal policies. To address the issues of mechanism heterogeneity and related data scarcity, we propose a data-efficient RL algorithm that exploits structural causal models (SCMs) to model the state dynamics,  which are estimated by leveraging both commonalities and differences across subjects. The learned SCM enables us to counterfactually reason what would have happened had another treatment been taken. It helps avoid real (possibly risky) exploration and mitigates the issue that limited experiences lead to biased policies. We propose counterfactual RL algorithms to learn both population-level and individual-level policies. We show that counterfactual outcomes are identifiable under mild conditions and that $Q$-learning on the counterfactual-based augmented data set converges to the optimal value function. Experimental results on synthetic and real-world data demonstrate the efficacy of the proposed approach. 
\blfootnote{\footnotesize $^1$University of Cambridge, $^2$Carnegie Mellon University, $^3$Max Planck Institute for Intelligent Systems, $^4$Alan Turing Institute, $^5$Imperial College London, $^*$Equal Contribution. Correspondence Authors:  \texttt{cl641@cam.ac.uk, biweih@andrew.cmu.edu}.}
\end{abstract}

\section{Introduction}
Over the last few years, reinforcement learning (RL) has been successfully applied to challenging problems such as playing Go \cite{AlphaGo} and Atari games \cite{Atari}. Key factors of the success include a substantial amount of interaction data and that the ongoing tasks are well-designed in a fixed environment. However, these factors do not always hold in real-world scenarios. A case in point is the RL formulation of healthcare, which aims to optimize sequential treatments to achieve recovery \cite{RL_health2, RL_health3}. Healthcare data usually have the properties that (1) only a few records are available for each patient and no further exploration can be performed, and that (2) patients may show different responses to identical treatments. These two properties impede the application of most RL algorithms to learn optimal policies.
 
Regarding handling the issue of data scarcity, model-based approaches tend to be more sample efficient and allow better interpretability than model-free ones. However, model-based approaches have problems when dealing with complex dynamics. Several hybrid approaches have been considered to mitigate these limitations. For example, model-based value expansion (MVE \cite{MVE}) incorporates a fixed depth of predictive models of system dynamics into model-free value function estimation, and later stochastic ensemble value expansion (STEVE \cite{STEVE}) further mitigates the bias due to model mis-specification by dynamically interpolating various horizon lengths. Besides, a recent approach based on counterfactually-guided policy search \cite{CounterRL1} models the dynamics with a pre-defined structural causal model (SCM) and performs probabilistic counterfactual reasoning to generate alternative outcomes under counterfactual actions. Its implementation assumes that the ground-truth transition and reward kernels are all given, which may not be realistic in some cases. 

Regarding heterogeneity across subjects or environments, a common strategy is to use meta-RL. For instance, context-based meta-RL methods adapt to new tasks by aggregating experience into a latent representation on which the policy is conditioned \cite{Meta_RL2, Meta_RL3, Meta_RL4, Meta_RL1}. In practice, meta-RL is usually hard to train, and during training, it may require large amounts of data drawn from a large set of distinct tasks, exacerbating the problem of sample efficiency.

To address the issues of mechanism heterogeneity and related data scarcity, we propose a sample-efficient RL algorithm, leveraging the following properties:
\begin{enumerate}
	\item Although the treatment effect may be different across individuals, a large proportion may still show a similar trend. We leverage commonalities and take into account variations across individuals to achieve more reliable estimation. 
	\item We leverage structural causal models (SCMs) to model the dynamic process. For generality, we do not put hard constraints on the functional class of causal mechanisms or data distributions. Moreover, to take into account mechanism heterogeneity, we include a variable $\theta_C$ in the causal system explicitly, to characterize hidden factors which change across individuals. 
	\item Given the SCM, accordingly we perform counterfactual reasoning by following Pearl's procedure \cite{Pearl00}. Counterfactual reasoning is the process of evaluating conditional claims about alternate possibilities and their consequences. For example, in our case, we can leverage it to infer that given the observed state-action tuple $\langle S_t=s_t,A_t=a,S_{t+1}=s_{t+1} \rangle$, what would have happened had we performed $a'$. This helps avoid real (possibly risky) exploration, which is often infeasible in real-world scenarios, and mitigates the problem that limited experiences lead to biased policies. 
\end{enumerate}

\textbf{Contributions}: We propose a practically useful and theoretically sound approach to tackling mechanism heterogeneity and data scarcity in RL, by counterfactual data augmentation. 1) On the technical side, we achieve flexible counterfactual reasoning in the general case, by adopting neural networks for function approximation of the SCM with no hard restrictions on data distribution or causal mechanisms, and use it for data augmentation to address the issues in RL. 2) We explicitly model the changing factors across subjects to achieve personalized policies, as well as a general policy over the population. 3) Theoretically, we show that the counterfactual outcome is identifiable under rather weak conditions. Furthermore, we show that $Q$-learning on the counterfactual-based augmented data set converges to the optimal value function. 

\section{Preliminaries} \label{Sec: Preliminaries}

In this section, we briefly introduce structural causal models and counterfactual reasoning which will be used throughout the paper. 

\subsection{Structural Causal Models}

Let $\mathbf{Y} = \{Y_1, \cdots, Y_n\}$ be a set of $n$ observed variables. A structural causal model consists of a set of equations of the form \cite{Pearl00}:
\begin{equation}
Y_i = f_i(X_i,U_i), 
\label{equ:SCM}
\end{equation}
for $i = 1,\cdots,n$, where $X_i$ stands for the set of parents of $Y_i$, i.e., a subset of the remaining variables in $\mathbf{Y}$ that directly cause $Y_i$, and $U_i$ represents disturbances (noises) due to omitted factors. Each of the functions $f_i$ represents a causal mechanism that determines the value of $Y_i$ from the causes and the noise term on the right side. The functional characterization in Eq. (\ref{equ:SCM}) provides a convenient language for specifying how the resulting distribution would change in response to interventions.

\subsection{Counterfactual Reasoning}
\label{Sec: counterfactual}

Suppose we performed action $a$ in state $s_t$ and observed $s_{t+1}$. One may be interested in knowing what would have happened had we performed $a'$. This is a \textit{counterfactual} question. It has been shown that given an SCM, we can then perform counterfactual reasoning \cite{Pearl00}. 

Suppose the SCM in (\ref{equ:SCM}) is given, denoted by $M$, and that we have evidence $Y=y$ and $X=x$ (the subscript $i$ has been omitted for representation convenience). The following steps show how to counterfactually infer $Y$ had we set $X=x'$ \cite{Pearl00}:
\begin{itemize}
	\item Step 1 (abduction): Use evidence $(Y=y,X=x)$ to determine the value of $U$.
	\item Step 2 (action): Modify the model, $M$, by removing the structural equations for the variables in $X$ and replacing them with the function $X=x'$, to obtain the modified model, $M_{x'}$.
	\item Step 3 (prediction): Use the modified model, $M_{x'}$, and the value of $U$ to compute the counterfactual value of $Y$.
\end{itemize}
The counterfactual outcome is usually represented as: $Y_{X=x'}|Y=y,X=x$. Note that in Step 1, we perform deterministic counterfactual, that is, counterfactuals pertaining to a single unit of the population, where the value of $U$ is determined.

\section{Counterfactual RL Using SCMs} \label{Sec: CTRL}

We now propose data-efficient RL algorithms, by leveraging SCMs and its corresponding counterfactual-based data augmentation to handle the issues of data scarcity and mechanism heterogeneity. In this section, we first propose CounTerfactual Reinforcement Learning of a general policy, denoted by CTRL$_g$, aiming for a policy over the population. Next, we propose CounTerfactual Reinforcement Learning of personalized policies, denoted by CTRL$_p$, providing personalized policies for each individual or each automatically determined group.

\subsection{\texorpdfstring{\textbf{CTRL}$_{g}$}{}: Estimation of a General Policy } \label{Sec: General}

When examining whether a treatment is effective and should be adopted as a standard, it is first essential to understand its effect over the population. In the following, we focus on the estimation of a general policy for the population.

We assume that the state $S_{t+1}$ satisfy the SCM
\begin{equation}
S_{t+1} = f(S_t, A_t, U_{t+1}),
\label{FCM}
\end{equation}
where $f$ represents the causal mechanism, $A_t$ the action at time $t$, and $U_{t+1}$ the noise term, which is independent of $(S_t; A_t)$. To estimate a general policy, we do not consider the variability across individuals.

Given observed triplets $\langle S_t, A_t, S_{t+1} \rangle$ from individuals, for $t = 1,\cdots, T$, the first problem is how to efficiently estimate the causal mechanism $f$. To achieve generality, we do not specify a particular functional class of the causal mechanism, e.g., linear relations \cite{Shimizu06}, nonlinear relations with additive noise \cite{Hoyer09}, or the causal model with further post-nonlinear transformations \cite{Zhang_UAI09}. Instead, we use a generative adversarial framework to learn $f$, by minimizing the discrepancy between real data and generated data. In addition, to achieve counterfactual-based data augmentation, we need to estimate the value of the noise term at each time point in (\ref{FCM}). 

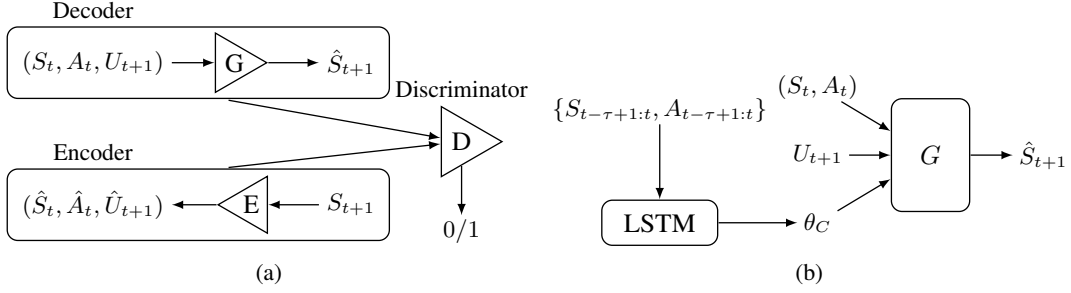
\begin{figure}
\centering
 \begin{subfigure}[b]{.5\textwidth}
 \centering
   \begin{tikzpicture}[boxx1/.style={draw,rounded corners,minimum height=0.7cm,text width=2cm,align=center,text centered}, boxx2/.style={draw,rounded corners,minimum height=0.95cm,text width=5cm,align=center,text centered}, tria1/.style={draw,regular polygon,regular polygon sides=3,shape border rotate = 90,minimum height=0.01cm,text width=0.2cm,align=center,text centered}, tria2/.style={draw,regular polygon,regular polygon sides=3,shape border rotate = 270,minimum height=0.01cm,text width=0.2cm,align=center,text centered}, tria3/.style={draw,regular polygon,regular polygon sides=3,shape border rotate = 270,minimum height=0.15cm,text width=0.3cm,align=center,text centered}, scale=0.7, line width=0.5pt, inner sep=0.4mm, shorten >=.1pt, shorten <=.1pt]

			\draw (-1,0) node(1) [boxx2, draw] {};
			\draw (-3,0) node(2)  {{\footnotesize\,$(S_t,A_t, U_{t+1})$\,}};
			\draw (1.9,0) node(3)  {{\footnotesize\,$\hat{S}_{t+1}$\,}};
			\draw (-0.35,0) node(4) [tria2, draw] {G};
			\draw (4,-1.5) node(6) [tria3, draw] {D}; 
			\draw (-1,-2.7) node(8) [boxx2, draw] {};
			\draw (-3,-2.7) node(9)  {{\footnotesize\,$(\hat{S}_t,\hat{A}_t,\hat{U}_{t+1})$\,}};
			\draw (1.9,-2.7) node(10)  {{\footnotesize\,$S_{t+1}$\,}};
			\draw (0,-2.7) node(11) [tria1, draw] {E};
			
            \draw (-.5,-0.7) node(14){};
            \draw (-.5,-2) node(15){};

			\draw (4,-3.2) node(16)  {{\footnotesize\,$0/1$\,}};
			
			\draw (-3,1) node(17)  {{\footnotesize\,Decoder\,}};	
			\draw (4,-0.5) node(18)  {{\footnotesize\,Discriminator\,}};	
			\draw (-3,-1.7) node(19)  {{\footnotesize\,Encoder\,}};	
			
			\draw[-latex] (2) -- (4); 
			\draw[-latex] (4) -- (3); 
			\draw[-latex] (14) -- (6);
			\draw[-latex] (10) -- (11); 
			\draw[-latex] (11) -- (9); 
			
			\draw[-latex] (6) -- (16); 
			\draw[-latex] (15) -- (6); 
		\end{tikzpicture}
		\caption{} \label{fig:BGAN}
	\end{subfigure}%
	~~
	\begin{subfigure}[b]{0.5\textwidth}
	 \centering
			\begin{tikzpicture}[boxx1/.style={draw,rounded corners,minimum height=1.5cm,text width=1cm,align=center,text centered}, boxx2/.style={draw,rounded corners,minimum height=0.6cm,text width=1.5cm,align=center,text centered}, scale=0.6, line width=0.5pt, inner sep=0.2mm, shorten >=.1pt, shorten <=.1pt]
			\draw (8.5,-1.5) node(1) [boxx1, draw] {$G$};
			\draw (6, 0) node(3)  {{\footnotesize\,$(S_t,A_t)$\,}};
			\draw (6, -1.5) node(4)  {{\footnotesize\,$U_{t+1}$\,}};
			\draw (6,-3.0) node(5) {{\footnotesize\,$\theta_{C}$\,}};
			\draw (11, -1.5) node(6)  {{\footnotesize\,$\hat{S}_{t+1}$\,}};
			\draw (2.5,-3.0) node(7) [boxx2, draw] {LSTM};
			\draw (2.5,-0.5) node(8) {{\footnotesize\,$\{S_{t-\tau+1:t}, A_{t-\tau+1:t}\}$\,}};
			
			\draw[-latex] (3) -- (1); 
			\draw[-latex] (4) -- (1); 
			\draw[-latex] (5) -- (1); 
			\draw[-latex] (1) -- (6); 
			\draw[-latex] (7) -- (5); 
			\draw[-latex] (8) -- (7); 	
			\end{tikzpicture}
			\caption{}
			\label{fig:LSTM_BGAN}
		\end{subfigure} 
		\caption{\footnotesize (a) Generator $G$, Encoder $E$, and Discriminator $D$ in CTRL$_g$.  (b) Generator in CTRL$_p$.}
		\label{fig:BiCoGAN} 
\end{figure}

To achieve the two goals, estimating $f$ and noise values simultaneously, we cast the learning of both an inference machine (encoder) and a deep generative model (decoder) in a generative adversarial network (GAN)-like adversarial framework, called Bidirectional Conditional GAN (BiCoGAN \cite{BiCoGAN}). Specifically, the BiCoGAN contains two parts: one is a generative model mapping from $\langle S_t, A_t, U_{t+1} \rangle$ to $S_{t+1}$, and the other is an inference mapping from $S_{t+1}$ to $\langle S_t, A_t, U_{t+1} \rangle$. The discriminator is trained to discriminate between joint samples from encoder distribution
 \begin{equation*}
     P(S_{t+1}, \hat{S}_t, \hat{A}_t, \hat{U}_{t+1}) = P(S_{t+1})P(\hat{S}_t, \hat{A}_t, \hat{U}_{t+1}|S_{t+1})
 \end{equation*}
 and decoder distribution
 \begin{equation*}
	\! \!P(\hat{S}_{t+1},\! S_t,\! A_t,\! U_{t+1}) \!  = \!P(S_t,\! A_t, \!U_{t+1})P(\hat{S}_{t+1}|S_t,\! A_t,\! U_{t+1}),
  \end{equation*}
 where $\hat{S}$, $\hat{A}_t$, and $\hat{U}_{t+1}$ denote the estimations of $S$, $A_t$, and $U_{t+1}$, respectively. The decoder and the encoder learn conditionals $P(\hat{S}_{t+1}|S_t, A_t, U_{t+1})$ and $P(\hat{S}_t, \hat{A}_t, \hat{U}_{t+1}|S_{t+1})$, respectively, to fool the discriminator, with objective function:
  \begin{small}\begin{equation*}
	\centering
	\begin{array}{ll}
	& \min\limits_{G,E} \max\limits_{D} V(D,G,E) =
	 \min\limits_{G,E} \max\limits_{D} \big\{ \underbrace{\mathbb{E}_{S_{t+1} \sim P_{\text{data}}(S_{t+1})} [\log D(E(S_{t+1}),S_{t+1})]}_{\text{Encoder}} \\ 
	& \qquad + \underbrace{\mathbb{E}_{\ddot{Z}_t \sim P_{\ddot{Z}_t}(\ddot{Z}_t)} [\log( 1- D(G(\ddot{Z}_t),\ddot{Z}_t))]}_{\text{Decoder}} 
	+ \underbrace{\lambda \mathbb{E}_{(S_t,A_t,S_{t+1}) \sim P_{\text{data}}(S_t,A_t,S_{t+1})}[R((S_t,A_t),E(S_{t+1})]}_{\text{Regularizer}} \big \},
	\end{array}
\end{equation*}\end{small}where $\ddot{Z}_t = (S_t, A_t, U_{t+1})$, $D$ denotes a discriminator network, $G$ a generator network, $E$ an encoder network, and $R$ is a regularizer (with hyperparameter $\lambda$) to prevent overfitting the estimation error of $(S_t, A_t)$, which helps to better encode extrinsic factors. A graph illustration is given in Figure \ref{fig:BGAN}.

After learning the SCM, including the causal mechanism $\hat{f}$ and noise values $\hat{u}_{t+1}$, for $t=1,\cdots,T$, we can counterfactually reason what would have happened if another treatment had been taken. Suppose at time $t+1$, we have $\langle S_t=s_t,A_t=a,S_{t+1}=s_{t+1} \rangle$. We  want to know what would have been the next state, if we had taken the action $a'$. Practically, this can be achieved by feeding $s_t$, $a'$, and $\hat{u}_{t+1}$ into the learned generator network $G$, and the output is the counterfactual outcome $s'_{t+1}$.
	
The alternative actions $a'$ are chosen in the following way. Suppose the action space is $\mathcal{A}\sim P_{\mathcal{A}}$ with support $[b^-,b^+]$, which can be either discrete or continuous; for example, in healthcare, whether surgery is performed is a binary variable, while a drug dosage may be continuous. We uniformly sample from $[b^-,b^+]$ to generate alternative actions $a'$ and accordingly estimate the counterfactual outcome $s'_{t+1}$.  We denote by $\tilde{\mathbb{D}}$ the augmented data set after including the data from counterfactual reasoning. 
Counterfactually reasoning about the effects of alternative actions helps avoid possibly risky exploration and mitigates the problem that limited experience leads to biased policies in RL. 

\paragraph{Remark 1 (SCMs vs. standard model-based approaches).} 
SCMs directly involve noise term $U$, which makes counterfactual reasoning possible. In principle, standard model-based approaches (e.g., \cite{PILCO, MDRL1}) do not have $U$ as input and cannot directly produce counterfactual outcomes--counterfactual reasoning is not possible without a causal model. Moreover, our approach nicely benefits from the power of the recently developed GAN-type methods in capturing the property of high-dimensional (conditional) distributions and generating new random samples. 
By contrast, the dynamics in standard model-based approaches are usually parameterized with a Gaussian distribution or a Gaussian mixture distribution. Although in the limit of infinitely many components, Gaussian mixture can fit any distribution, this becomes unpractical if the numbers get too large.

\paragraph{Remark 2 (causal reasoning vs. standard Monte-Carlo simulations).}
Standard Monte-Carlo simulations are not able to directly perform (individual-level) counterfactual reasoning, and hence cannot achieve our goal. As pointed out by Pearl \cite{Pearl00}, in counterfactual reasoning one has to exploit a causal model, and then he/she is able to infer the specific property of the considered individual, related to the noise term $U$, and then exploit it to derive what would have happened if we had performed alternative action $a’$ (for the same individual). Monte-Carlo simulations can only produce random samples at the population (not individual) level.

\subsection{\texorpdfstring{\textbf{CTRL}$_{p}$}{}: Estimation of Personalized Policies}\label{Sec: Specific}

In healthcare, patients can exhibit different responses to identical treatments. Therefore, one should care not only about the treatment effect for the general population, but also the response of each individual or each properly divided group. In the following, we propose counterfactual RL of personalized policies (CTRL$_p$), by taking into account variabilities across individuals/groups and leveraging commonalities to achieve statistically reliable estimation.
	
We use a variable $\theta_{C}$ to explicitly take into account hidden factors, whose value may vary across individuals, where $C$ denotes the subject index. Thus, for an individual, we assume that the state $S_{t+1}$ satisfies the SCM
\begin{equation}
	S_{t+1} = f(S_t, A_t, \theta_{C}, U_{t+1}),
	\label{FCM_C}
\end{equation}
where $f$ represents the overall family of mechanisms, and $\theta_C$ captures factors that depends on the subject or group.

To capture the variation (i.e., estimate the value of $\theta_{C}$), we segment the data sequence of each subject using sliding windows of size $\tau$, resulting in triplets $
\{\langle S_{t-i+1}, A_{t-i+1}, S_{t-i+2} \rangle\}_{i=1}^{\tau}$. 
At each time $t$, we exploit individual-specific information from the sequence $\{S_{t-\tau+1:t}, A_{t-\tau+1:t}\}$, by leveraging Long-Short Term Memory (LSTM) \cite{LSTM}. The LSTM output $\hat{\theta}_{C}$ acts as a new input to the generator $G$. Note that we constrain the same individual to have the same value for $\theta_{C}$. Figure \ref{fig:LSTM_BGAN} illustrates the generator $G$.
	
Similar to CTRL$_g$, CTRL$_p$ is also estimated with BiCoGAN, with the difference that there is a latent variable $\theta_{C}$, learned with an LSTM network, as a new conditioning variable, i.e., conditioning variables $\ddot{Z} = (S_t, A_t, \theta_{C},U_{t+1})$.
All parameters in CTRL$_p$ are learned simultaneously in an adversarial manner.

After learning the SCM, including $f$, $\theta_{C}$, and  $U_{t+1}$, we divide individuals into groups by applying k-means clustering to the estimated values of $\hat{\theta}_{C}$. We use the estimated centroids from k-means as a new $\tilde{\theta}_C$, and hence $\tilde{\theta}_C$ is constant within each group but varies across groups. We can then perform counterfactual reasoning on each group of individuals, as described in Section \ref{Sec: General}, resulting in an augmented dataset $\check{\mathbb{D}}_i$ for the $i$-th group.

\subsection{Identifiability}

Given triplets $\langle S_t=s_t, A_t=a, S_{t+1}=s_{t+1} \rangle$, it is important to show whether the counterfactual outcome is identifiable. 
Without this guarantee, the output of the method may be different from the true counterfactual outcome. Surprisingly, the following theorem shows that without any hard constraints on the functional form $f$ and the noise distribution in the SCM, the derived counterfactual outcome is correct under weak assumptions. This makes
counterfactual reasoning generally possible.
	
\begin{Theorem}
		Suppose $S_{t+1}$ satisfies the following structural causal model: 
		 \begin{equation*}
		   S_{t+1} = f(S_t, A_t, U_{t+1}),
		 \end{equation*}
		 where $ U_{t+1} \independent (S_t; A_t)$, and we assume that $f$ (which is unknown) is smooth and strictly monotonic in $U_{t+1}$ for fixed values of $S_t$ and $A_t$. Suppose we have observed $\langle S_t = s_t, A_t = a, S_{t+1}  = s_{t+1} \rangle$. Then for the counterfactual action $A_t = a'$, the counterfactual outcome
		 \begin{equation}
		  S_{{t+1}, A_t = a'} | S_t = s_t, A_t = a, S_{t+1}  = s_{t+1}
		 \end{equation}
		 is identifiable.
		 \label{Theorem: identifiability}
\end{Theorem}

Note that the above theorem naturally holds for the specific SCM in Eq. (\ref{FCM_C}), since for an individual $C=c$, $\theta_c$ is fixed and thus, equivalently, Eq. (\ref{FCM_C}) can be written as $S_{t+1} = f_c(S_t, A_t, U_{t+1})$.
The monotonicity condition on $f$ with respect to $U_{t+1}$ guarantees that the noise term is recoverable. Consider extreme cases, nonlinear causal models with additive noise or multiplicative noise, where the effect variable is always strictly monotonically increasing in the noise, rendering the noise recoverable from the cause and effect. Moreover, the above theorem holds no matter whether the state space and the action space are continuous or discrete.

Furthermore, note that although the function $f$ and the probabilistic distribution $P(U_{t+1})$ are not uniquely identifiable from the collected data without strong constraints on the functional class of $f$ and on the data distributions \cite{zhang2016estimation}, surprisingly, the constructed counterfactual outcome does not depend on which $f$ and $P(U_{t+1})$ we choose, provided that $f$ is strictly monotonic in $U_{t+1}$. In experiments, this strict monotonicity can be easily implemented through monotonic multi-layer perceptron network \cite{lang2005monotonic}, in which positive signs of the weights are guaranteed by introducing their exponential form \cite{zhang1999feedforward}.

\paragraph{Remark 3.} 
Previously, the identifiability of counterfactual quantities was only shown in the case where both cause $X$ and effect $Y$ are binary and $Y$ is monotonic relative to $X$ \cite{Pearl00}. Later, \cite{oberst2019counterfactual} discusses the case with categorical variables. Note that in previous work, the monotonicity is with respect to $X$ (equivalent to $A_t$ in Eq.(\ref{FCM})), while in our theorem, it is relative to the noise term $U$. Interestingly, we find that by requiring monotonicity in $U$, we can show identifiability for more general data types and causal mechanisms. Moreover, note that here we only care about the identifiability of the counterfactual outcomes, which does not require identifiability of the SCM.

\begin{algorithm}[t]
   	\caption{Policy Learning via Counterfactual-Based Data Augmentation}
   	\begin{enumerate}
     \item \textbf{Input:} observed triplets $\langle S_t, A_t, S_{t+1} \rangle$ from individuals, for $t = 1,\cdots, T$.
     \item \textbf{Estimation of a general policy:}
      \begin{itemize}
          \item[2.1.] Estimate the SCM given in (\ref{FCM}) with BiCoGAN.
          \item[2.2.] Generate counterfactual data given alternative actions, according to the estimated SCM. Denote the counterfactually augmented data set by $\tilde{\mathbb{D}}$.
          \item[2.3.] Perform D3QN learning on $\tilde{\mathbb{D}}$.
      \end{itemize}
     \item \textbf{Estimation of personalized policies:}
      \begin{itemize}
          \item[3.1.] Estimate the SCM given in (\ref{FCM_C}) with BiCoGAN, meanwhile with LSTM to characterize individual specific factor $\theta_C$.
          \item[3.2.] For each individual or each automatically determined group, generate counterfactual data given alternative actions, according to its estimated SCM. Denote the counterfactually augmented data set for the $i$-th individual by $\check{\mathbb{D}}_i$.
          \item[3.2.] Perform D3QN learning on $\check{\mathbb{D}}_i$.
      \end{itemize}
    \item \textbf{Output:} General policy over the population and personalized policies for each individual or each group.  
   	\end{enumerate}
   	\label{Alg} 
\end{algorithm}

\section{Deep Q-Networks} \label{Sec: DeepQ}

After estimating the dynamics model and generating an augmented data set $\tilde{\mathbb{D}}$ by counterfactual reasoning, we are now ready to learn policies on $\tilde{\mathbb{D}}$ to maximize future rewards. To this end, we use the Dueling Double-Deep Q-Network (D3QN) \cite{RL_sepsis1}, a variant of Deep Q-Networks (DQNs) \cite{Deep_Q1}. 
	
A simple DQN may suffer from shortcomings that Q-values are often overestimated, which leads to inaccurate predictions and sub-optimal policies \cite{RL_sepsis1}. This problem can be mitigated by using Double-Deep Q-Networks \cite{DoubleQ}, where the target Q values are determined using actions found through a feed-forward pass on the main network, instead of being estimated directly from the target network. 
	
In addition, a high Q value may be due to (1) a patient's positive state, e.g., being near discharge, or (2) a treatment that is taken at that time step \cite{RL_sepsis1}. For general applicability of the learned policy, it is important to distinguish between these two cases. To achieve this, one may use Dueling Q-Networks \cite{DuelingQ}, where $\mathcal{Q}(s,a)$ is split into separate value and advantage streams, representing the quality of the current state and the quality of the chosen action, respectively. Leveraging the advantages from Double-Deep Q-Networks and Dueling Q-Networks, the final network is a fully connected Dueling Double-Deep Q-Network. Algorithm \ref{Alg} gives the entire procedure of policy learning via counterfactual-based data augmentation. 

The following theorem shows that on the counterfactually augmented data set, $Q$-learning converges to the optimal value function.
\begin{Theorem}
   Given the transition dynamics, $Q$-learning on the counterfactually augmented data set converges with probability one to the optimal value function $Q^*$, as long as the state and action spaces are finite, and the learning rate $\alpha_t$ satisfies $\sum_t \alpha_t = \infty$ and $\sum_t \alpha_t^2 < \infty$.
 \label{Theorem: converge}
\end{Theorem} 

\paragraph{Remark 4.} 
In this paper, we used deterministic counterfactuals (see Section \ref{Sec: counterfactual}), where the value of $U_{t+1}$ can be determined, with which we show that the estimation is consistent. Instead, CF-PE \cite{CounterRL1} performs probabilistic counterfactuals, where the value of $U_{t+1}$ is sampled from the posterior distribution $P(U_{t+1}|\mathbb{D})$. CF-PE is unbiased, if there is no model mismatch, but it is not guaranteed to converge to the optimal value function. Moreover, if $U_{t+1}$ is directly sampled from the prior distribution $P(U_{t+1})$, which is usually the case in a probabilistic transition model, the estimation is even biased \cite{CounterRL1}.  

\section{Experimental Results}

To evaluate the proposed approaches CTRL$_g$ and CTRL$_p$, we applied them to a modified classical control problem and a real-world healthcare dataset. More experiments and details about model architectures and implementation are given in Appendix. 

\subsection{Results on Classical Control Problems}

We first evaluated the performance of our proposed methods on the cartpole environment in OpenAI gym, a benchmark for classical control tasks. It has continuous states with dimension $d_s = 4$ and discrete actions with dimension $d_a = 1$. To increase the complexity and make the environment stochastic, we extended the original two discrete actions (i.e., $a=0, 1$) to eleven actions (i.e., $a=0, 0.1,\ldots, 0.9, 1$)  and added 5\% Gaussian noise to both states and actions. 

To train our models and evaluate their performance, we created two datasets: \textbf{SD}: Simple dataset with fixed gravity ($g_{\text{earth}}=9.8$) and \textbf{HD}: Hybrid dataset with five different gravities ($g_{\text{jupiter}}=24.79$, $g_{\text{earth}}=9.8$, $g_{\text{mercury}}=3.7$, $g_{\text{neptune}}=11.15$, and $g_{\text{pluto}}=0.62$). In \textbf{SD}, we collected 250 trials by applying random actions and put the collected data in five data subsets with the number of trials $n_\text{trial}= 50, 100, ..., 250$. Each trial has 20 consecutive steps and each step contains $\langle s_t, a_t, s_{t+1}, r_t \rangle$. In \textbf{HD}, for each gravity, we generated 50 trials in the same way as that in \textbf{SD}. Additionally, we generated 5-time-step sequential data by applying sliding windows along each trajectory for $\text{CTRL}_{p}$.

\paragraph{Comparisons to Baselines and SOTA}
We first compared the proposed $\text{CTRL}_{g}$ with three well-known baselines and three state-of-the-art (SOTA) approaches in terms of sample efficiency on \textbf{SD}. The three baselines are summarised as follows (more details can be found in Appendix). (1) Base-$D$ (deterministic dynamics model): The next state $S_{t+1}$ is determined by current state $S_t$ and current action $A_t$ with $S_{t+1} = f(S_t,A_t)$. (2) Base-$S$ (probabilistic dynamics model with unimodal distribution): The probabilistic dynamics model $P(S_{t+1}|S_t,A_t;\theta)$ is Gaussian, where $\theta$ denotes a set of parameters in the dynamics. (3) Base-$M$ (probabilistic dynamics model with multimodal distribution): $P(S_{t+1}|S_t,A_t;\theta)$ is a mixture of Gaussians, implemented by a mixture density network \cite{MDN_Bishop}. The three SOTA approaches are STEVE \cite{STEVE}, BCQ \cite{fujimoto2018off}, and PETS \cite{MDRL1}.

\begin{figure}
\centering
 \begin{subfigure}[b]{.4\textwidth}
  \centering
\includegraphics[width=.9\linewidth]{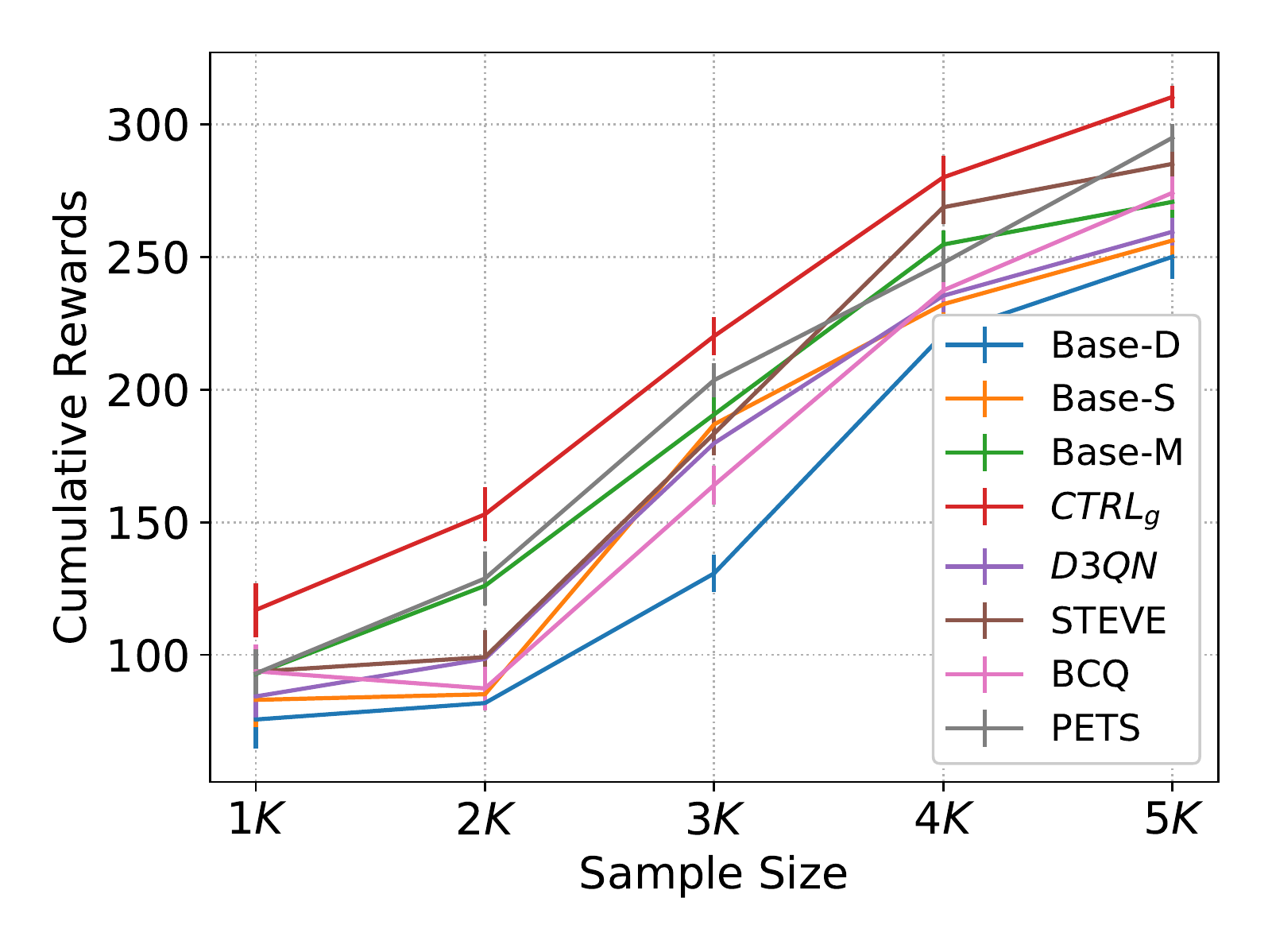}
  \caption{}
  \label{fig:cart_sd}
 \end{subfigure}%
 \begin{subfigure}[b]{.4\textwidth}
  \centering
\includegraphics[width=.9\linewidth]{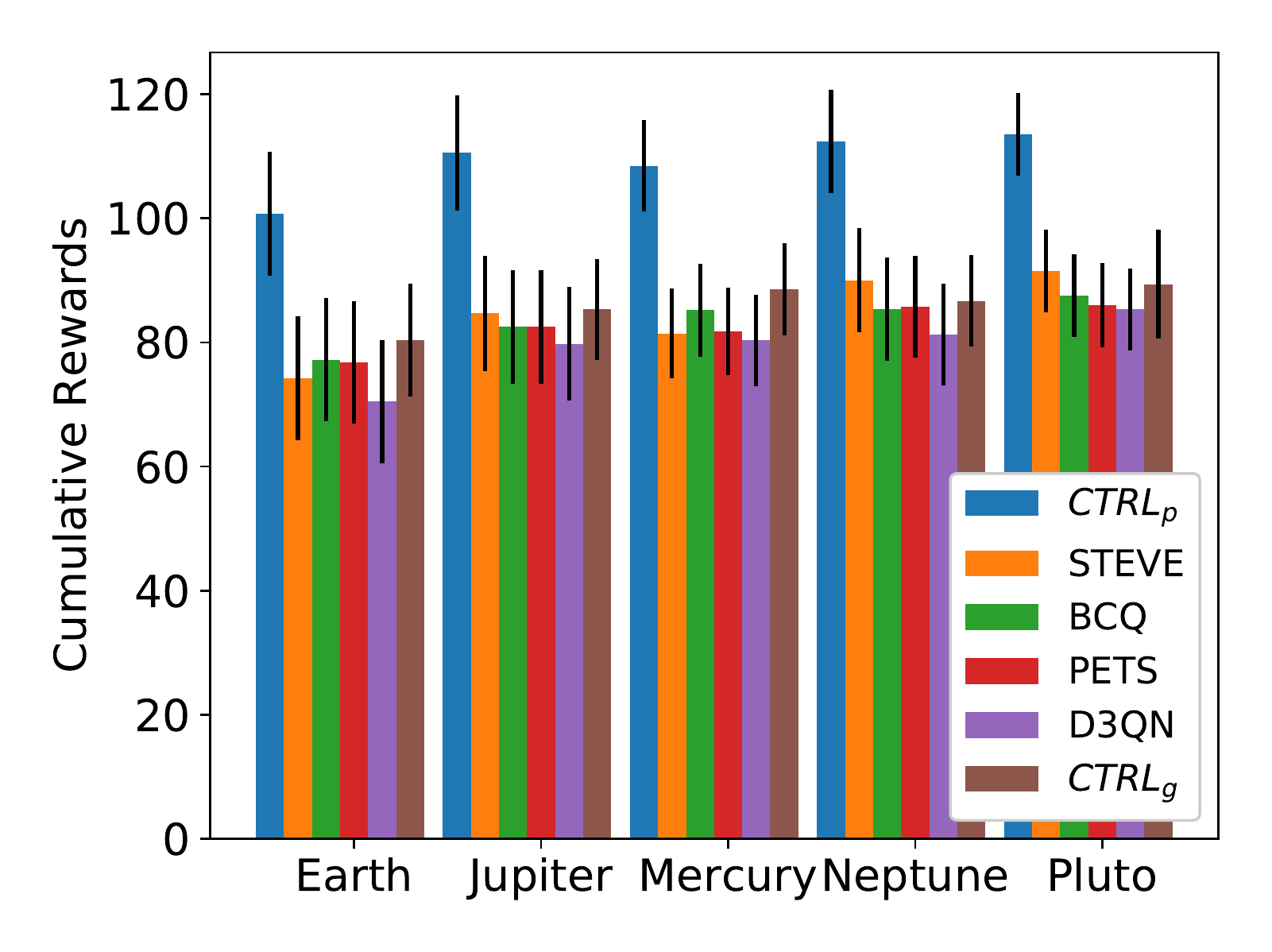}
  \caption{}
  \label{fig:cart_hd} 
 \end{subfigure}
 \caption{\footnotesize (a) $\text{CTRL}_{g}$ outperforms three baselines, three state-of-the-art methods, and D3QN on the original data on \textbf{SD}. The improvement is obvious, especially when the sample size is small.  (b) The general policy $\text{CTRL}_{g}$ is  competitive, and the personalized policy $\text{CTRL}_{p}$ outperforms state-of-the-art methods on \textbf{HD}.}
\end{figure} 

\begin{figure*}[t] 
  \centering
    \includegraphics[width=\linewidth]{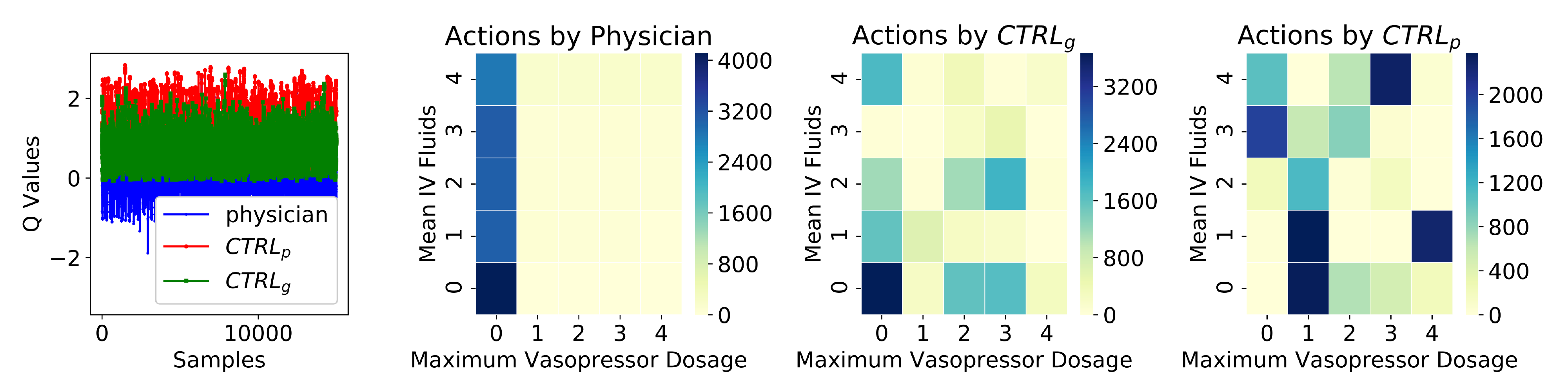}
    \text{(a)\qquad\qquad\qquad\quad\quad(b)\qquad\qquad\qquad\quad\quad(c)\quad\qquad\qquad\qquad\qquad(d)} 
  \caption{\footnotesize (a) Comparison of Q-values estimated by Physician, $\text{CTRL}_{g}$, and $\text{CTRL}_{p}$ over all state-action pairs on the test set. (b-d) Distribution of actions obtained by applying Physician, $\text{CTRL}_{g}$, and $\text{CTRL}_{p}$. The axes labels index the discretized action space, where 0 represents no drug given, and 4 is the maximum dosage.} 
  \label{fig:sepsis_ctrl} 
\end{figure*}

For a fair comparison, all methods are trained on the five subsets in \textbf{SD} (with $n_\text{trial}= 50, 100, ..., 250$) without exploration in the environment, and all results are reported after training converges. After learning the transition dynamics, we counterfactually generated new data for each and trained a D3QN on a mixture of real data and generated data. For the baseline models, counterfactual reasoning was replaced by generating $s_{t+1}$ via fixing $s_t$ and applying a random $a_t$. Then we tested the learned policy in the CartPole simulation by performing 10 trials with different random seeds. We also compare with D3QN on the original data, without data augmentation, denoted by D3QN. Figure \ref{fig:cart_sd} shows that $\text{CTRL}_{g}$ outperforms all other methods, with the highest cumulative reward in all settings, especially when the sample size is small. 

Next, we evaluated the performance of $\text{CTRL}_{p}$ on \textbf{HD}. Based on the grouping information obtained by $\text{CTRL}_{p}$, we counterfactually generated more data for each group and then learned an optimal policy on each. At test time, we first estimated the group membership of a given patient, and then applied the optimal group-specific policy. Figure \ref{fig:cart_hd} demonstrates that the individual policy learned by $\text{CTRL}_{p}$ is superior to the general policy learned by the SOTA methods in all five environments, as shown by the consistently highest cumulative rewards. Note that the SOTA methods learn a general policy, and only our method is individualized. We did not compare with meta-RL, because typically meta-RL-based approaches are not in the batch off-policy setting.

\subsection{Results on MIMIC-III}

In this section, we investigated the performance of our approach on the real-world {\em Medical Information Mart for Intensive Care-III} (MIMIC-III) database with Sepsis-3 \cite{johnson2016mimic}. Each patient state consists of 46 variables including laboratory tests, vital signs, and patient demographics. We follow the setting of data preparation and the reward function in \cite{RL_sepsis1}. We selected 15,656 patients who have at least five consecutive time steps. For each model, we generated the train, validation, and test sets. More precisely, $\text{CTRL}_{g}$ had three sets of size, 120K, 15K, and 15K, respectively. Since $\text{CTRL}_{p}$ needs the sequences as input, the sizes of the three sets were smaller, corresponding to 80K, 10K, and 10K.

\paragraph{General Treatment with $\textbf{CTRL}_{g}$}
We first learned SCM on the (real) training set. Once the model was learned, we randomly selected a pair of $(s_t, s_{t+1})$ from the set, and used the learned encoder to estimate the noise value $\hat{u}_t$. We then used $s_t$ and $\hat{u}_t$ to generate $\hat{s}_{t+1}$ by taking a uniformly random action. After training the D3QN on the augmented data set, we estimated the Q-values of the physician policy as well as that of the learned optimal policy (Q-values have been typically used to evaluate patient care, e.g., \cite{RL_sepsis1, RL_health2, RL_health3}). In Figure \ref{fig:sepsis_ctrl}\textcolor{blue}{a} we can see that the optimal policy trained on a combination of real data and the generated data by $\text{CTRL}_{g}$ achieves a larger estimated Q-value than the physician policy. In other words, the optimal policy trained on our counterfactually-generated data is able to increase survival time. Figure \ref{fig:sepsis_ctrl}\textcolor{blue}{(b-c)} further shows that the policy learned by $\text{CTRL}_{g}$ differs from the physician policy. While both polices exhibit a similar trend to high usage of IV fluids and lower usage of vasopressors, $\text{CTRL}_{g}$ recommends using more vasopressors and less IV fluids.

\paragraph{Personalized Treatments with $\textbf{CTRL}_{p}$}

We trained $\text{CTRL}_{p}$ on sequential data with five time steps. During training, our model uses k-means on the estimated $\theta_{C}$ to group the patients.\footnote{More sophisticated clustering methods may lead to improved results \cite{schulam2015clustering}.} Exploiting this grouping information, we pool the data from the patients within the same group, then counterfactually generate new data based on the real data in each group, and finally train a D3QN on a mix of real data and generated data for each group. The resulting optimal policy is specific for each group. During the test phase, given a new patient with a sequence of data, we first estimated which group he or she belongs to, and then applied the corresponding policy of that group. As shown in Figure \ref{fig:sepsis_ctrl}\textcolor{blue}{a}, applying a personalized policy to each patient leads to a larger average estimated Q-value than directly applying the general policy learned from the whole population data. Figures \ref{fig:sepsis_ctrl}\textcolor{blue}{(b-d)} present the difference of action statistics over all clusters by applying different policies. It seems that compared to the general policy, the optimal personalized policy on average prefers heavier usage of both vassopressors and IV fluids. Interestingly, Figure \ref{fig:sepsis_ctrl}\textcolor{blue}{(c-d)} suggest that actions learned by $\text{CTRL}_{p}$ are more diversely distributed than that by $\text{CTRL}_{g}$. This indeed makes sense in that the personalized policy encourages personalized treatments for each patient, which can lead to more diversity. It is also reasonable that the policy learned by $\text{CTRL}_{g}$ is closer to the physician policy than that by $\text{CTRL}_{p}$, because the physician policy also comes from their experience of treatment to the population as $\text{CTRL}_{g}$ does. Although without a real-world experiment we cannot verify the true effect of the learned optimal policy, it may provide guidance for physicians to inform their decisions and do further evaluations.

\section{Related Work}

Many studies have been trying to address the perceived sample inefficiency of RL. Compared to model-free RL, model-based RL tends to be more sample efficient and allow better interpretability. However, current model-based RL algorithms may converge to suboptimal solutions \cite{MBRL_bias}. The reason is that the learned models may fail to reflect the true process from insufficient interaction data and thus lead to biased policies. One way to mitigate this problem is to  incorporate uncertainty into the dynamics model. For example, the deterministic dynamics model has been extended by parameterizing it with a Gaussian distribution or a Gaussian mixture distribution for better generalization \cite{MDN_Bishop}. PILCO \cite{PILCO} leverages Gaussian processes to express model uncertainty, and further the model uncertainty is incorporated into planning and policy evaluation. Probabilistic Ensembles with Trajectory Sampling algorithm (PETS) \cite{MDRL1} combine uncertainty-aware deep network dynamics models with sampling-based uncertainty propagation. 

Recently, an algorithm based on probabilistic counter\-factually-guided policy search has been proposed in the POMDP setting \cite{CounterRL1}. In its implementation, it assumes that the ground-truth transition, observation and reward kernels are all given. Furthermore, it adds uncertainty only on the initial state $S_1$. Although noise added to the initial state can propagate to the whole system, it is not the true causal process. In contrast, we model and implement the underlying causal process with an SCM in the MDP setting, so the noise appears in the SCM at each time step (see Eq.~(\ref{FCM})), representing unmeasured factors influencing $S_{t+1}$. In addition, the work by \cite{Data_fusion_17} leverages counterfactual data in multi-armed bandit problems under heterogeneous conditions, with counterfactual quantities being estimated by active agents empirically. Backtracking models \cite{backtracking_model} consider predicting the samples that may lead to high-reward states.

\section{Conclusion}

To address the issue of dynamics heterogeneity and data scarcity common in healthcare, we propose a data-efficient RL algorithm that exploits SCMs to model the state dynamics. The learned SCM enables us to counterfactually reason what would happen had another treatment been taken. It helps avoid real (possibly risky) exploration and mitigates the problem that limited experience leads to biased policies. We provide both a general policy over the population and personalized policies for individuals in automatically identified groups. The proposed methods show promising results on both synthetic and real-world datasets. 

\newpage

\bibliographystyle{unsrt}
\bibliography{ms}

\newpage

\appendix

\section{Proof of Theorem 1}

\begin{proof}
  It is clear that as implied by Lemma 1 of \cite{zhang2016estimation}, the function $f$ and the probabilistic distribution $P(U_{t+1})$ are not uniquely identifiable from the collected data. Furthermore, from the triplets $\langle S_t = s_t,A_t = a, S_{t+1} = s_{t+1}\rangle$, one can always find a infinite number solutions functions to $f$ and $P(U_{t+1})$ such that $S_{t+1} = f(S_t;A_t;U_{t+1})$, where $U_{t+1} \independent (S_t;A_t)$ and $f$ is strictly monotonic in $U_{t+1}$. Choose an arbitrary solution to $f$ and $P(U_{t+1})$, denoted by $f^i$ and $P^i(U_{t+1})$. Later, surprisingly, we will see that the constructed counterfactual outcome actually does not depend on the index $i$; that is, it is independent of which $f^i$ and $P^i(U_{t+1})$ we choose.
  
  Given observed evidence $\langle S_t = s_t,A_t = a, S_{t+1} = s_{t+1} \rangle$, because $f^i$ is strictly monotonic in $U_{t+1}^i$, we can determine $\hat{u}_{t+1}^i$, which is the value of $U_{t+1}^i$, with
  \begin{equation*}
     \hat{u}_{t+1}^i = {f_{s_t,a}^i}^{-1}(s_{t+1}),
  \end{equation*}
  where ${f_{s_t,a}^i}^{-1}$ denotes the inverse of $f^i$ with $S_t = s_t,A_t = a$ fixed. Then, we can determine the value of the cumulative distribution function of $U_{t+1}^i$ at $\hat{u}_{t+1}^i$, denoted by $\alpha^i$. 
  
  Without loss of generality, we first show the case where $f^i$ is strictly increasing in $U_{t+1}^i$. Because $f^i$ is strictly increasing in $U_{t+1}^i$ and $s_{t+1} = f^i(s_t, a, \hat{u}_{t+1}^i)$, $s_{t+1}$ is the $\alpha^i$-quantile of $P(S_{t+1}| S_t=s_t, A_t=a)$. Then it is obvious that since $s_{t+1}$ and $P(S_{t+1} | S_t=s_t, A_t=a)$ are determined, the value of $\alpha_i$ is independent of the index $i$, that is, it is identifiable. Thus, below, we will use $\alpha$, instead of $\alpha_i$. 
  
  Since $U_{t+1} \independent (S_t;A_t)$, when doing interventions on $A_t$, the value $\hat{u}_{t+1}^i$ will not change. The counterfactual outcome $S_{t+1, A_t=a'} |S_t = s_t,A_t = a, S_{t+1} = s_{t+1}$ can be calculated as follows:
  \begin{equation*}
      S_{t+1, A_t=a'} = f^i(S_t = s_t, A_t = a',\hat{u}_{t+1}^i).
  \end{equation*}
  Because $\hat{u}_{t+1}^i$ does not change after the intervention, the counterfactual outcome $S_{t+1, A_t=a'} |S_t = s_t,A_t = a, S_{t+1} = s_{t+1}$ is the $\alpha$-quantile of the conditional distribution $P(S_{t+1} | S_t = s_t, A_t = a')$. This quantile exists and it depends only on the conditional distribution $P(S_{t+1} | S_t = s_t, A_t = a')$, but not the chosen function $f^i$ and $P^i(U_{t+1})$, rendering the counterfactual outcome identifiable.

  Similarly, the above reasoning procedure can also be applied to the case where $f^i$ is strictly decreasing in $U_{t+1}^i$.
  
  Therefore, the counterfactual outcome is identifiable, under the condition that $f$ is strictly monotonic in $U_{t+1}$.
\end{proof}

\section{Proof of Theorem 2}
	\begin{proof}
	  From the proof of Theorem 1, we can see that the counterfactual outcome $S_{t+1, A_t=a'} |S_t = s_t,A_t = a, S_{t+1} = s_{t+1}$ is the $\alpha$-quantile of the conditional distribution $P(S_{t+1} | S_t = s_t, A_t = a')$. Given the transition dynamics $P(S_{t+1} | S_t, A_t)$, the counterfactually augmented triplet can be determined, and it satisfies the underlying Markov decision process (MDP). Moreover, it has been shown that Q-learning on the data generated from the MDP converges to the optimal value function, under the listed conditions \cite{Convergence_Qlearning, Convergence_Qlearning2}. Therefore, $Q$-learning on the counterfactually augmented data set converges to the optimal value function $Q^*$.
	\end{proof}

\section{Three Baselines}
Below, we give more details about the three baselines, Base-$D$, Base-$S$, and Base-$M$, which all belong to model-based RL.

Model-based RL explicitly builds a model of the state transition and evaluates actions by searching this model. It is appealing because it is sample efficient. 
According to the dynamics model that is used, it can be divided into two types: model-based RL with deterministic dynamics models and that with probabilistic ones. Therefore, we compared the proposed $\text{CTRL}_{g}$ and $\text{CTRL}_{p}$ with three well-known baselines in terms of sample efficiency.
	
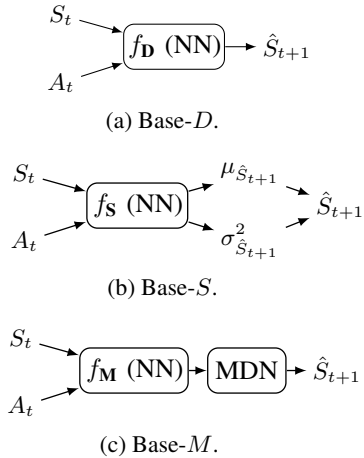
\begin{figure}[htp!]
		  \centering	
		  		\begin{subfigure}{0.23\textwidth}
			\centering
			\begin{tikzpicture}[boxx1/.style={draw,rounded corners,minimum height=0.6cm,text width=1.3cm,align=center,text centered}, scale=0.6, line width=0.4pt, inner sep=0.2mm, shorten >=.1pt, shorten <=.1pt]
			\draw (2.5,-0.7) node(1) [boxx1, draw] {$f_{\textbf{D}}$ (NN)};
			\draw (0, 0) node(2)  {{\footnotesize\,$S_t$\,}};
			\draw (0, -1.5) node(3)  {{\footnotesize\,$A_t$\,}};
			\draw (5, -0.7) node(6)  {{\footnotesize\,$\hat{S}_{t+1}$\,}};
			
			\draw[-latex] (2) -- (1); 
			\draw[-latex] (3) -- (1); 
			\draw[-latex] (1) -- (6); 
			
			\end{tikzpicture}
			\caption{Base-$D$.}
		\end{subfigure}%
		\vspace{.3cm}
		\\
		\begin{subfigure}{0.3\textwidth}
			\centering
			\begin{tikzpicture}[boxx1/.style={draw,rounded corners,minimum height=0.6cm,text width=1.3cm,align=center,text centered}, scale=0.6, line width=0.5pt, inner sep=0.2mm, shorten >=.1pt, shorten <=.1pt]
			\draw (2.5,-0.7) node(1) [boxx1, draw] {$f_{\textbf{S}}$ (NN)};
			\draw (0, 0) node(2)  {{\footnotesize\,$S_t$\,}};
			\draw (0, -1.5) node(3)  {{\footnotesize\,$A_t$\,}};
			\draw (5, -0.0) node(6)  {{\footnotesize\,$\mu_{\hat{S}_{t+1}}$\,}};
			\draw (5, -1.5) node(7)  {{\footnotesize\,$\sigma^2_{\hat{S}_{t+1}}$\,}};
			\draw (7, -0.7) node(8)  {{\footnotesize\,$\hat{S}_{t+1}$\,}};
			
			\draw[-latex] (2) -- (1); 
			\draw[-latex] (3) -- (1); 
			\draw[-latex] (1) -- (6); 
			\draw[-latex] (1) -- (7);
			\draw[-latex] (6) -- (8);
			\draw[-latex] (7) -- (8);
			
			\end{tikzpicture}
			\caption{Base-$S$.}
		\end{subfigure}%
		\vspace{.3cm}
		\\
		\begin{subfigure}{0.3\textwidth}
			\centering
			\begin{tikzpicture}[boxx1/.style={draw,rounded corners,minimum height=0.6cm,text width=1.4cm,align=center,text centered}, boxx2/.style={draw,rounded corners,minimum height=0.6cm,text width=1cm,align=center,text centered}, scale=0.6, line width=0.5pt, inner sep=0.2mm, shorten >=.1pt, shorten <=.1pt]
			\draw (2.5,-0.7) node(1) [boxx1, draw] {$f_{\textbf{M}}$ (NN)};
			\draw (0, 0) node(2)  {{\footnotesize\,$S_t$\,}};
			\draw (0, -1.5) node(3)  {{\footnotesize\,$A_t$\,}};
			\draw (5, -0.7) node(6)  [boxx2, draw] {MDN};
			\draw (7, -0.7) node(8)  {{\footnotesize\,$\hat{S}_{t+1}$\,}};
			
			\draw[-latex] (2) -- (1); 
			\draw[-latex] (3) -- (1); 
			\draw[-latex] (1) -- (6); 
			\draw[-latex] (6) -- (8); 
			
			\end{tikzpicture}
			\caption{Base-$M$.}
		\end{subfigure}
		\vspace{.3cm}
	   \caption{Three baselines: (a) Base-$D$, (b) Base-$S$, and (c) Base-$M$.}
		\label{fig:base}
	\end{figure}

Specifically, for model-based RL with a deterministic dynamics model, the next state $S_{t+1}$ is determined by current state $S_t$ and current action $A_t$: $S_{t+1} = f(S_t,A_t)$, where $f$ can be learned with neural networks. Figure \ref{fig:base}(a) gives a graphical illustration of the generating process with a deterministic dynamics model; we denote it by Base-$D$. It has been observed that model-based RL with deterministic dynamic models tends to converge to sub-optimal solutions where the learned dynamics may not reflect the true process. One way to mitigate this problem is to properly incorporate uncertainty into the dynamics model.
	
For model-based RL with probabilistic dynamics, one can represent the conditional distribution of the next state given the current state and action in  parameterized form, $P(S_{t+1}|S_t,A_t;\theta)$, where $\theta$ denotes a set of parameters in the dynamics that needs to be learned. One may parameterize the dynamics model with a Gaussian distribution or a mixture distribution \cite{MDN_Bishop} for better generalization. Figure \ref{fig:base}(b) illustrates a generating process of mean $\mu_{\hat{S}_{t+1}}$ and variance $\sigma^2_{\hat{S}_{t+1}}$ of a Gaussian distribution, and $\hat{S}_{t+1}$ is sampled from the learned Gaussian distribution; we denote it by Base-$S$. Figure \ref{fig:base}(c), instead, uses a mixture density network (MDN), to handle more general cases; we denote it by Base-$M$. 
	
\section{More Experimental Results}	

\subsection{More Results on Classical Control Problems}
\begin{figure}[htp!]
  \centering
\includegraphics[width=.6\linewidth]{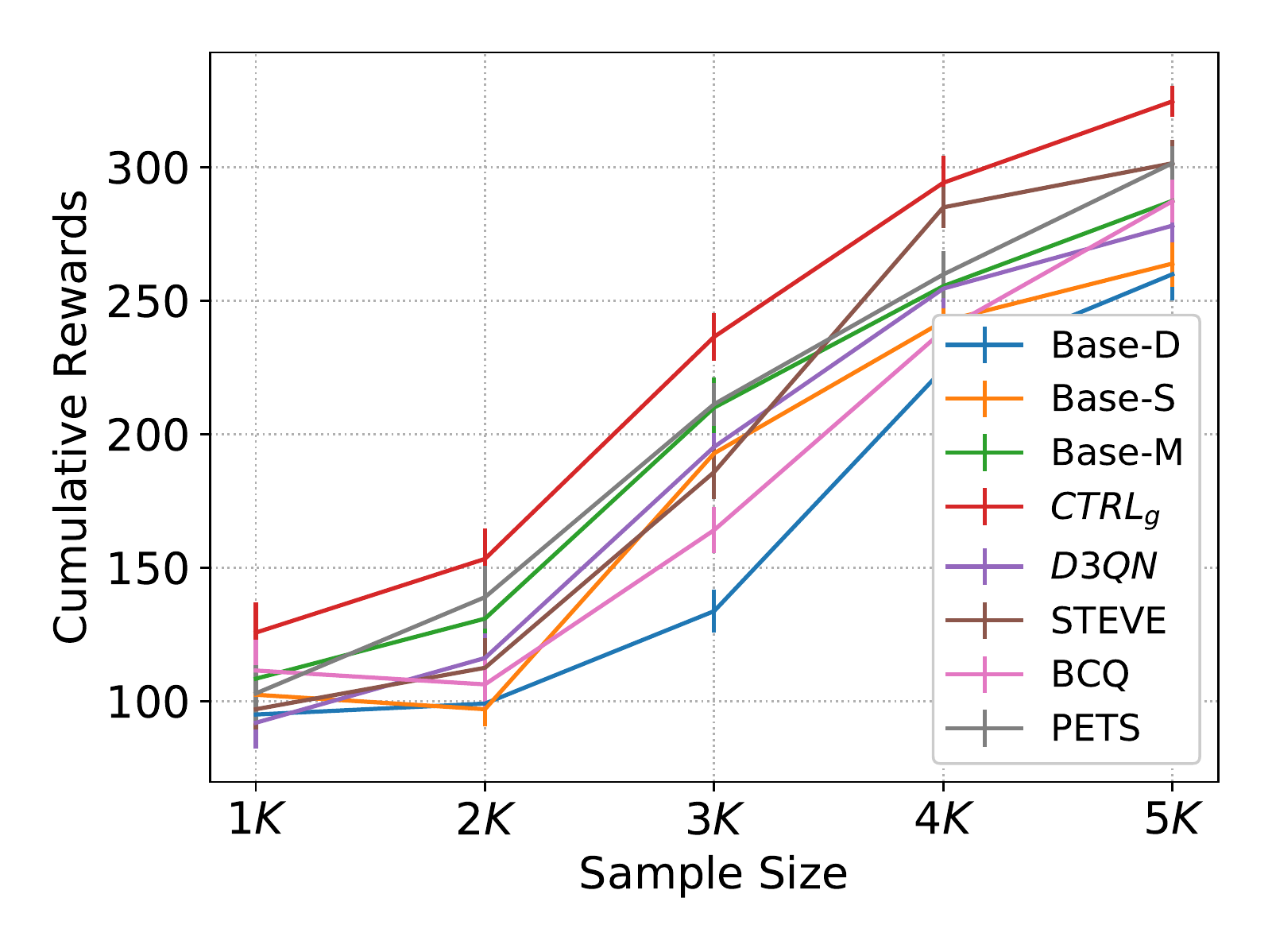} \vspace{-0.2cm}
  \caption{\footnotesize Comparison of \text{CTRL}$_{g}$, \text{CTRL}$_{p}$, three baselines, and three state-of-the-art methods on \textbf{SD}.}
  \label{fig:cart_nonrand_sd}
\end{figure}

\begin{figure}[htp!]
  \centering
\includegraphics[width=.6\linewidth]{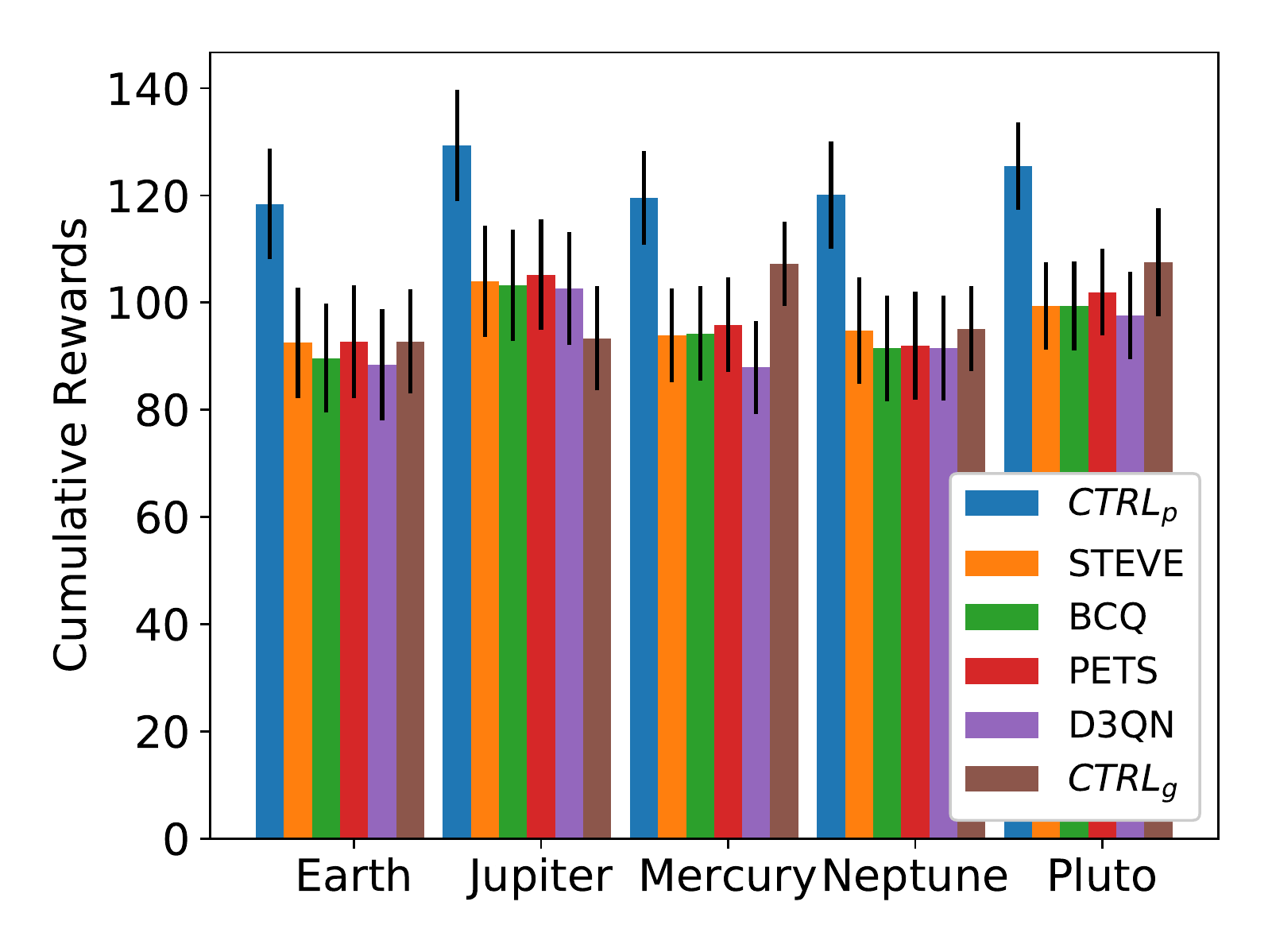} \vspace{-0.2cm}
  \caption{\footnotesize Comparison of \text{CTRL}$_{g}$, \text{CTRL}$_{p}$, and three state-of-the-art methods on \textbf{HD}.}
  \label{fig:cart_nonrand_hd}
\end{figure}

We conduct additional experiments on the data generated under a non-random policy which is a better-than-random policy trained after ten episodes, as shown in Figures \ref{fig:cart_nonrand_sd} and \ref{fig:cart_nonrand_hd}.

\section{Network Architectures}

In all experiments, unless stated otherwise, we used linear layers, followed by batch normalization and nonlinear activation function (ReLU) in all network architectures. For simplicity, we omit the notation of batch normalization and ReLU in the following description. Note that, in the implementation of CTRL algorithms, the strict monotonicity stated in \textbf{Theorem 1} are easily implemented through monotonic multi-layer perceptron network \cite{lang2005monotonic}, in which positive signs of the weights are guaranteed by introducing their exponential form \cite{zhang1999feedforward}.

\subsection{Network Architecture for Baseline Models} 
Here we give details about network architectures of the three baseline models. The network structure is presented in Table \ref{table1}.

\begin{table}[htp!]
\centering
\caption{Network structures of baseline models}\vspace{2mm}
\label{table1}
\begin{tabular}{ccc}
\hline
 & Hidden Layers & Neurons Per Layer   \\
\hline        
Base-$D$ & 2             & 300              \\ 
Base-$S$ & 2             & 300                   \\ 
Base-$M$ & 2             & 300             \\  
\hline
\end{tabular}
\end{table}

Note that, in the model of Base-$M$, the hyperparameter $\alpha$, which represents the categorical probability of the mixture density network, is set to 5.

\subsection{Network Architecture for \texorpdfstring{\text{CTRL}$_{g}$}{}} 
Shown in Table \ref{tab:ctrl1}.

\begin{table}[htp!]
\centering
\footnotesize
\caption{Network structures of \text{CTRL}$_{g}$}\vspace{2mm}
\label{tab:ctrl1}
\begin{tabular}{ccc}
\hline
 & Hidden Layers & Neurons Per Layer   \\
\hline        
Generator & 4            & 200 $\rightarrow$ 400 $\rightarrow$ 600 $\rightarrow$ 600               \\ 
Encoder & 4             & 600 $\rightarrow$ 600 $\rightarrow$ 400 $\rightarrow$ 200                     \\ 
Discriminator & 4             & 600 $\rightarrow$ 600 $\rightarrow$ 400 $\rightarrow$ 200                                \\  
\hline
\end{tabular}
\end{table}

\subsection{Network Architecture for \texorpdfstring{\text{CTRL}$_{p}$}{}} 

Shown in Table \ref{tab:ctrl2}.

\begin{table}[htp!]
\centering
\footnotesize
\caption{Network structures of \text{CTRL}$_{p}$}\vspace{2mm}
\label{tab:ctrl2}
\begin{tabular}{ccc}
\hline
 & Hidden Layers & Neurons Per Layer   \\
\hline        
Generator & 4            & 200 $\rightarrow$ 400 $\rightarrow$ 600 $\rightarrow$ 600               \\ 
Encoder & 4             & 600 $\rightarrow$ 600 $\rightarrow$ 400 $\rightarrow$ 200                     \\ 
Discriminator & 4             & 600 $\rightarrow$ 600 $\rightarrow$ 400 $\rightarrow$ 200                                \\  
LSTM & 1 & 200 \\
\hline
\end{tabular}
\end{table}

\subsection{Network Architecture for D3QN} 

Shown in Table \ref{tab:ddqn}.

\begin{table}[htp!]
\centering
\footnotesize
\caption{Network structures of DDQN}\vspace{2mm}
\label{tab:ddqn}
\begin{tabular}{ccc}
\hline
 & Hidden Layers & Neurons Per Layer   \\
\hline        
Policy & 4            & 512 $\rightarrow$ 512 $\rightarrow$ 512 $\rightarrow$ 512               \\ 
\hline
\end{tabular}
\end{table}

\end{document}